%% file: main.tex
\documentclass[preprint,12pt]{elsarticle}
\usepackage{multirow}  
\usepackage{booktabs}  
\usepackage{caption}   
\usepackage{graphicx} 
\usepackage{color} 
\usepackage{diagbox} 
\usepackage{subfigure} 
\usepackage{longtable}  
\usepackage{geometry}   
\usepackage[x11names,table,hyperref]{xcolor}
\usepackage[normalem]{ulem}
\usepackage{natbib}
\usepackage{hyperref}
\usepackage{setspace}

\usepackage{makecell}

\usepackage{amssymb}
\usepackage{amsmath}

\journal{Pattern Recognition}

\begin{document}
\doublespacing  

\begin{frontmatter}



\title{Instruction-Guided Fusion of Multi-Layer Visual Features in Large Vision-Language Models}

\author[label1]{Xu Li\fnref{co-first}}
\author[label2]{Yi Zheng \fnref{co-first}}
\author[label1]{Haotian Chen}
\author[label1]{Xiaolei Chen}
\author[label2]{Yuxuan Liang}
\author[label1]{Chenghang Lai}
\author[label2]{Bin Li}
\author[label1]{Xiangyang Xue\corref{corresponding}}

\fntext[equal]{The two authors contribute equally to this work.}
\cortext[corresponding]{Corresponding author.}

\affiliation[label2]{organization={School of Computer Science, Fudan University},
            addressline={220 Handan Road},
            city={Shanghai},
            postcode={200433},
            country={China}}
            
\affiliation[label1]{organization={Institute of Big Data, Fudan University},
            addressline={220 Handan Road},
            city={Shanghai},
            postcode={200433},
            country={China}}



\begin{abstract}
Large Vision-Language Models (LVLMs) have achieved remarkable success in a wide range of multimodal tasks by integrating pre-trained vision encoders and large language models. However, current LVLMs primarily rely on visual features extracted from the final layers of the vision encoder, overlooking the complementary information available in shallower layers. While recent approaches have explored the use of multi-layer visual features in LVLMs, they tend to be task-agnostic and fail to examine the dependencies of hierarchical visual features on specific tasks. To address these gaps, we systematically investigate the contributions of visual features from different encoder layers using 18 benchmarks spanning 6 task categories. Our findings reveal that multi-layer features provide complementary strengths with varying task dependencies, and uniform fusion leads to suboptimal performance. Building on these insights, we propose the instruction-guided vision aggregator, a module that dynamically integrates multi-layer visual features based on textual instructions, without increasing the number of visual tokens. Extensive evaluations demonstrate the superior performance of our method. Additionally, an in-depth analysis of the aggregator’s behavior highlights the dominance of mid-to-high-level features in semantic-rich tasks and the critical role of low-level features in fine-grained perception. This work provides valuable insights into the adaptive use of hierarchical visual features in LVLMs, offering a step toward more flexible and effective multimodal systems.
\end{abstract}




\end{frontmatter}



\input{introduction.tex} 
\input{relatedwork} 
\input{observations} 
\input{methods}
\input{experiments}

\input{conclusion}









\bibliography{references}
\end{document}

%% file: introduction.tex
\section{Introduction}
\label{sec1}

Recent advances in Large Language Models (LLMs) have fueled growing interest in equipping these models with visual capabilities. By integrating pre-trained vision encoders with LLMs, Large Vision-Language Models (LVLMs) have been developed, enabling joint interpretation of images and textual instructions \cite{llava1.5,instructblip,minigptv2}. These models combine the perceptual strengths of vision encoders with the reasoning capabilities of LLMs, achieving strong performance in diverse vision-language tasks with applications in fields such as medical diagnosis \cite{med}, autonomous driving \cite{driving}, and embodied intelligence \cite{embody}.

\begin{figure}[h!]
\centering
\includegraphics[width=\textwidth]{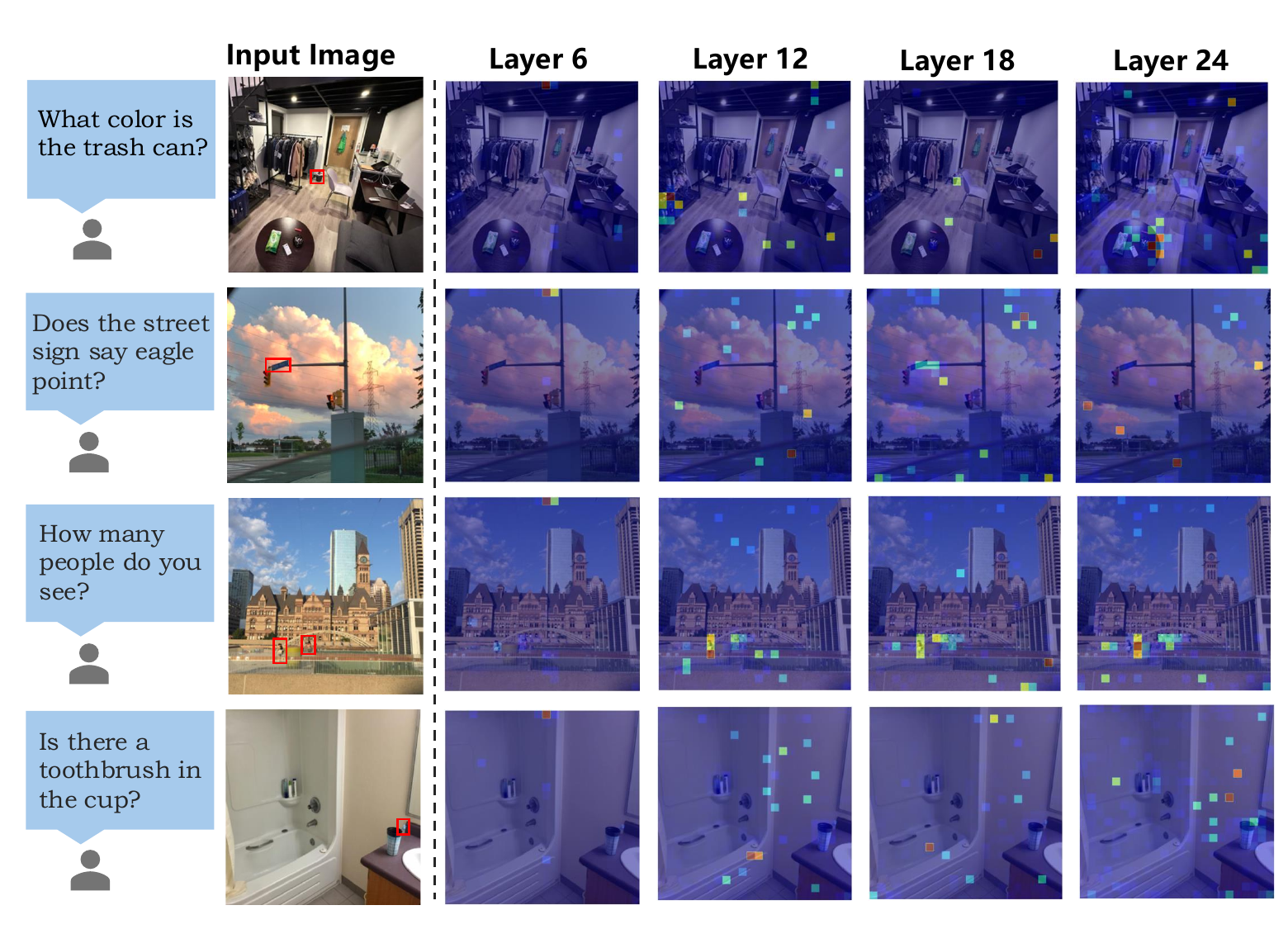}
\caption{Attention distribution of LVLMs across image patches when generating the answer token. Each column corresponds to a model trained with visual features extracted from a different layer of the vision encoder. Regions relevant to the text query are highlighted with red bounding boxes.} 
\label{fig:atten}
\end{figure}

The standard architecture of LVLMs consists of three primary components: a pre-trained vision encoder, a vision-language adapter, and a pre-trained LLM. The vision encoder extracts visual features from input images, which are subsequently aligned with the LLM's embedding space via the adapter. These aligned visual features are then processed together with the textual instruction by the LLM to generate a textual response. Typically, the vision encoder is implemented as a Vision Transformer (ViT) \cite{vit} that is pre-trained using image-text contrastive learning methods, such as CLIP \cite{clip} or SigLIP \cite{siglip}. While these encoders are adept at at capturing high-level semantic representations, they tend to overlook fine-grained visual details due to the limitations of thier web-scale training data, such as coarse or noisy text annotations and low-resolution images \cite{hallucination}. Consequently, LVLMs encounter challenges when tasked with problems requiring detailed visual understanding \cite{eyeswideshut}.

Recent research in visual representation learning highlights that low-level features in pre-trained vision encoders can capture detailed information, making them particularly effective for tasks requiring fine-grained visual perception \cite{teaching, vision}. Building on this, recent studies have explored the use of multi-layer visual features to enhance LVLM performance. DenseConnector \cite{denseconnector} integrates visual features from multiple encoder layers through downsampling and concatenation, while MMFuser \cite{mmfuser} uses high-level features as queries to extract visual details from shallower layers. Although both approaches improve performance, they fail to incorporate textual instructions during the fusion of hierarchical visual features. Given that textual instructions are critical in defining tasks for LVLMs, the inability to dynamically adjust feature fusion based on task-specific needs may result in suboptimal performance.

To the best of our knowledge, no prior work has systematically investigated how features from different layers of a vision encoder affect LVLM performance across various task categories. To address this, we trained LVLMs using features extracted from different layers of the vision encoder. As shown in Fig.~\ref{fig:atten}, the choice of visual feature layers directly influences the model’s focus on visual content, shaping its ability to process task-relevant information. Building on this observation, we conducted three-fold experiments in Section \ref{sec2}, which revealed the following key insights: 1) Visual features from different encoder layers enable LVLMs to excel in distinct downstream tasks; 2) Features from different layers are complementary, and their integration improves overall performance; 3) Static fusion of these features is suboptimal, while assigning different fusion weights leads to task-specific performance variations. 

\begin{figure}[h!]
\centering
\includegraphics[width=0.75\textwidth]{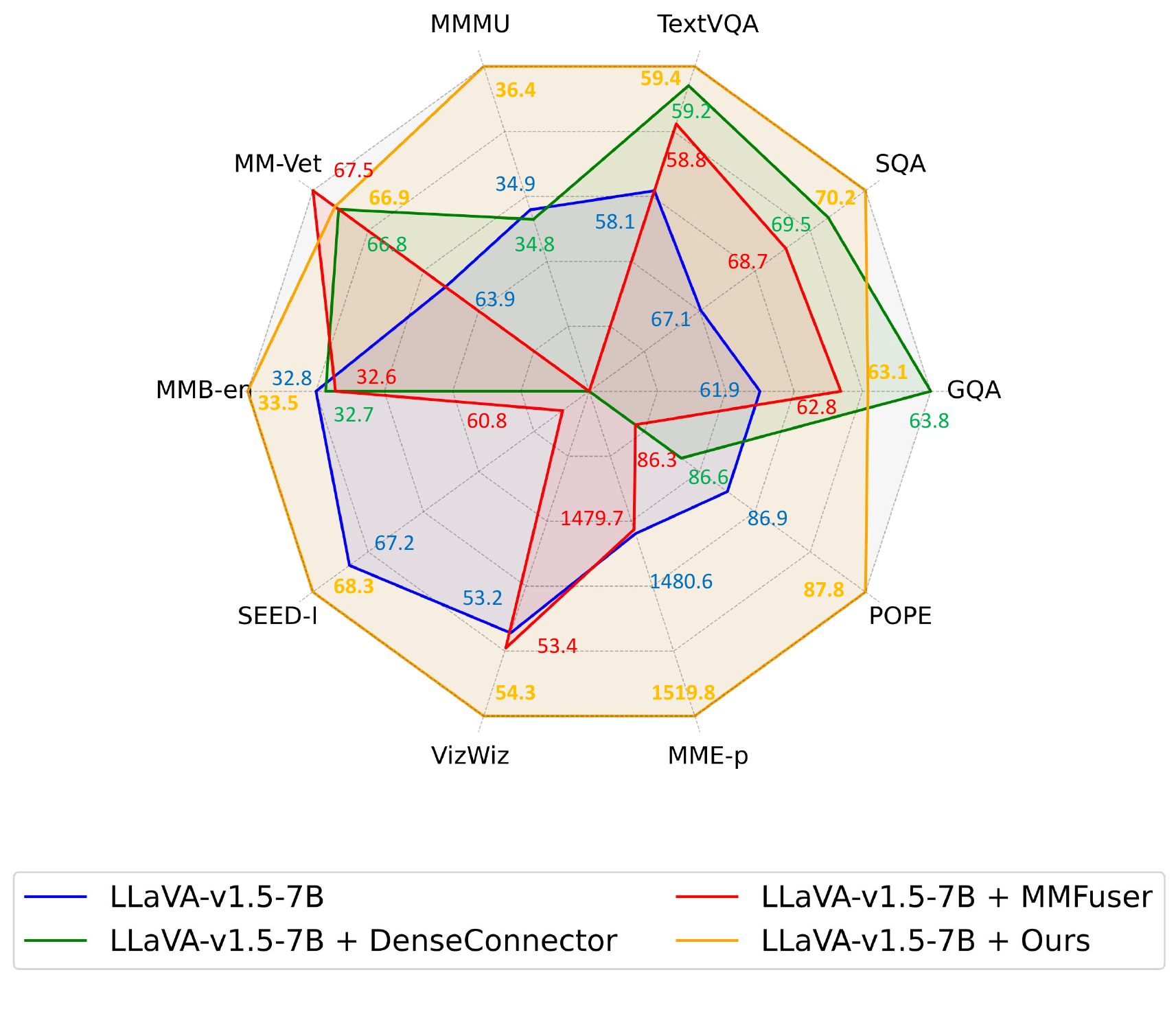}
\caption{Performance comparison of our method against the baseline model (LLaVA-v1.5-7B \cite{llava1.5}) and existing hierarchical visual feature fusion methods (DenseConnector \cite{denseconnector} and MMFuser \cite{mmfuser}) across 10 mainstream benchmarks.}
\label{fig:radarchart}
\end{figure}

Motivated by these insights, we posed the following question: can a lightweight module in LVLMs dynamically fuse visual features from different ecoder layers based on task-specific instructions? To answer this, we developed the instruction-guided vision aggregator within the typical LVLM framework. This module extracts task-relevant semantic information from input instructions and leverages it to assign fusion weights to hierarchical visual features, enabling the model to selectively emphasize the most relevant layers. When integrated into the widely-adopted LLaVA-v1.5 \cite{llava1.5}, it significantly improved baseline performance and outperformed existing task-agnostic fusion methods across diverse benchmarks, as shown in Fig~\ref{fig:radarchart}. 

In summary, our main contributions are as follows:

\begin{itemize}
    \item We conducted a comprehensive analysis of how visual features from different encoder layers affect LVLM performance across 18 benchmarks spanning 6 task categories, revealing the complementary nature of hierarchical features and the limitations of static fusion strategies.
  
    \item We proposed a lightweight module that dynamically allocates fusion weights to hierarchical visual features based on task-specific instructions, enabling task-driven feature integration without increasing the number of visual tokens.
    
    \item We integrated the proposed module into the LLaVA-v1.5 framework, achieving significant improvements over the baseline performance and surpassing existing task-agnostic multi-layer fusion methods as well as similarly scaled LVLMs.

    \item We analyzed the weight allocations of the proposed module, highlighting the dominance of mid-to-high-level features in semantic tasks and the critical role of low-level features in fine-grained perception, offering valuable insights into the adaptive use of hierarchical features in LVLMs.
\end{itemize}

%% file: relatedwork.tex
\section{Related Work}
\subsection{Large Vision-Language Models}
Recent breakthroughs in LLMs have enhanced language understanding and generation, which in turn has advanced vision-language integration, leading to the development of LVLMs such as LLaVA \cite{llava1.5}, MiniGPT-4 \cite{minigptv2}, and InstructBLIP \cite{instructblip}. These LVLMs typically consist of a vision encoder, a vision-language adapter, and an LLM. The vision encoder, usually based on a CLIP-ViT \cite{clip}, extracts visual features from input images. The vision-language adapter, commonly implemented as a multi-layer perceptron (MLP) \cite{llava1.5} or Q-former \cite{instructblip}, aligns these visual features to the LLM's input embedding space. Finally, the LLM generates responses by jointly processing the aligned visual features and the textual instructions.

Despite their successes, current LVLMs predominantly utilize features from the final or penultimate vision encoder layers, overlooking the rich visual details captured in shallower layers. Although effective in many tasks, this reliance limits their performance in scenarios requiring fine-grained visual understanding.

\subsection{Fusion of Multi-layer Visual Features in LVLMs}
Recent research has emphasized the distinct roles played by different layers in widely-adopted vision encoders. For instance, \cite{teaching} demonstrated that mid-to-high layers of CLIP-ViTs produce the most effective features for localized tasks, while global tasks predominantly benefit from features extracted from the final layers. Similarly, \cite{vision} highlighted the hierarchical nature of feature extraction in CLIP-ViTs: shallow layers capture basic textures, intermediate layers represent more complex patterns and object parts, and the deepest layers encode complete objects.

Building on these insights, several approaches have sought to integrate multi-layer visual features in LVLMs. \cite{denseconnector} proposed DenseConnector, which involves three strategies for multi-layer feature fusion: Sparse Token Integration, which downsamples features from pre-defined encoder layers and concatenates them with features from the final layer along the sequence dimension; Sparse Channel Integration, which concatenates features from pre-defined layers with those of the final layer along the channel dimension; and Dense Channel Integration, which groups features from all layers, applies downsampling within each group, and concatenates them along the channel dimension. Meanwhile, \cite{mmfuser} introduced MMFuser, which leverages semantically aligned deep features as queries to selectively extract key information from shallower layers, enriching the overall visual representation while preserving semantic coherence.

However, these methods do not account for task-specific instructions during feature fusion, lacking the ability to dynamically adjust fusion strategies in accordance with the demands of specific tasks. This limitation underscores the need for more adaptive methods that can optimize feature integration based on task characteristics.

\subsection{Text-guided dynamic fusion of visual features in LVLMs}
Text-guided feature fusion has emerged as a crucial strategy for improving task adaptation in LVLMs. \cite{instructblip} introduced an instruction-aware mechanism that incorporates instruction tokens as additional input to the Q-former, facilitating the extraction of task-relevant image features. \cite{move} proposed MoVE, which employs an instruction-based soft router to dynamically aggregate features from multiple vision encoders. Similarly, \cite{mova} introduced MoVA, a coarse-to-fine, text-guided approach that selectively integrates visual features from complementary vision encoders.

In contrast, our approach leverages a single vision encoder, dynamically assigning weights to hierarchical visual features based on textual instructions. This method enhances visual perception by prioritizing task-relevant features while avoiding architectural redundancy. 

%% file: observations.tex
\section{Preliminary Observations}\label{sec2}
Recent research underscores the significance of hierarchical visual features in LVLMs, but little attention has been paid to how features from different layers of the vision encoder impact LVLM performance across various downstream tasks. This gap can lead to suboptimal utilization of hierarchical visual information. To close the gap, we conducted a systematic analysis of the contributions of various encoder layers to LVLM performance across a wide range of task-specific benchmarks.

We adopted the architecture and training pipeline of LLaVA-v1.5-7B \cite{llava1.5} and utilized the evaluation set from \cite{cambrian}, which spans four primary task categories: General, Knowledge, Chart \& OCR, and Vision-Centric, covering a total of 15 benchmarks. To broaden the scope of evaluation, we introduce two additional task categories:

\begin{enumerate}
    \item \textbf{Fine-Grained}, which includes V*-bench \cite{vstar} and HR-bench \cite{hrbench} to test the model’s ability to perceive fine-grained details in high-resolution and complex visual scenes.
    \item \textbf{Object Hallucination}, which uses the POPE benchmark  \cite{pope} to evaluate the model’s susceptibility to object existence hallucination.
\end{enumerate}
With these additions, the final evaluation set comprises 6 task categories and 18 benchmarks.

\subsection{Layer-wise Performance Analysis}
We trained five models with varying encoder layer inputs: LLaVA-L6, LLaVA-L12, LLaVA-L18, LLaVA-L24, and LLaVA-L23 (a reproduction of LLaVA-v1.5 \cite{llava1.5}), each using the 6th, 12th, 18th, 24th, and 23rd (penultimate) encoder layers, respectively. These models were evaluated on the aforementioned benchmarks to understand how performance varied with different layers.

\begin{figure}[h!]
\centering
\includegraphics[width=\textwidth]{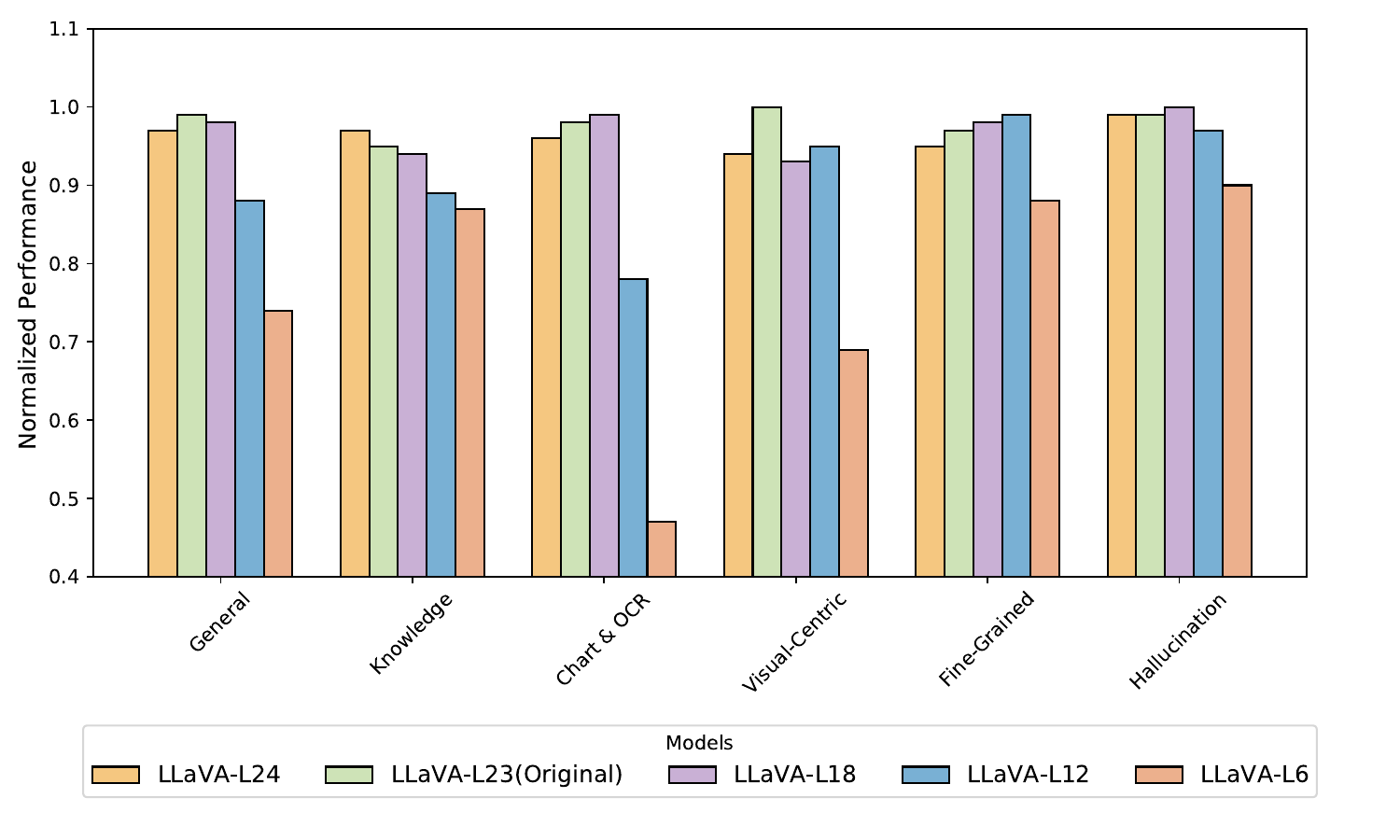}
\caption{Performance comparison of LVLMs trained using different single-layer visual features across 6 task categories. The vertical axis represents the normalized performance score, where each benchmark is scaled by setting the highest score to 1 and adjusting other models' scores accordingly.
}
\label{fig:performance}
\end{figure}

As shown in Fig~\ref{fig:performance}, the models exhibited distinct performance profiles across the task categories. Except for LLaVA-L6, all models excelled in at least one task category. Detailed results in Table~\ref{single-layer-visual-features} show that LLaVA-L23, utilizing the penultimate layer features, achieved the best overall performance, excelling in General and Vision-Centric tasks. This explains why penultimate layer features are commonly preferred in LVLMs. LLaVA-L24, leveraging the last layer, performed best on Knowledge tasks, while LLaVA-L18 outperformed others in Chart \& OCR and Object Hallucination tasks. Interestingly, LLaVA-L12, despite its lower overall ranking, scored the highest on Fine-Grained tasks, underlining the importance of low-level features in fine-grained visual perception. In contrast, LLaVA-L6, which utilized the relatively earliest encoder layer, struggled across all tasks due to its lack of semantic alignment. These results indicate that lower-level features are crucial for tasks requiring detailed visual perception, while higher-level features are better suited for tasks demanding global semantic understanding.

\begin{table*}[t!]
\caption{Performance comparison of LVLMs trained with different single-layer visual features across 18 benchmarks from 6 task categories. The overall score is calculated by normalizing the scores for each benchmark (setting the highest score to 1 and adjusting the others accordingly), followed by averaging the normalized scores across all benchmarks within each category. Bold values represent the best score in each row, while underlined values indicate the second-best score.}
\label{single-layer-visual-features}
\renewcommand{\arraystretch}{1}
\centering
\resizebox{\textwidth}{!}{  
\begin{tabular}{l|c|ccccc}
\toprule[2pt]
\multicolumn{2}{c|}{\diagbox[innerwidth=5.6cm,innerleftsep=5pt,innerrightsep=5pt]{Benchmarks}{Models}} &  LLaVA-L24     & LLaVA-L23       & LLaVA-L18     & LLaVA-L12     & LLaVA-L6 \\

\midrule[1pt]
\multirow{5}{*}{General}        & MME-p \cite{mme}        & \uline{1438.1}  & \textbf{1480.6} & 1390.4        & 1237.0        & 1021.1   \\
                                & GQA \cite{gqa}          & 61.7          & {\uline{61.9}}      & \textbf{63.1} & 60.3          & 53.8     \\
                                & SEED-I \cite{seed}       & 64.6          & {\uline{66.4}}      & \textbf{67.3} & 62.5          & 51.8     \\
                                & MMB-en \cite{mmb}      & 63.0          & \textbf{63.9}   & {\uline{62.9}}    & 53.3          & 41.8     \\
                                & Overall      & 0.97          & \textbf{0.99}   & {\uline{0.98}}    & 0.88          & 0.74     \\
\midrule[1pt]

\multirow{5}{*}{Knowledge}      & MathVista \cite{mathvista}    & {\uline{6.8}}     & \textbf{7.0}    & 6.5           & 5.6           & 5.8      \\
                                & MMMU \cite{mmmu}        & \textbf{36.3} & 34.9            & {\uline{35.6}}    & 34.8          & 33.3     \\
                                & AI2D \cite{ai2d}        & 50.0          & \textbf{55.2}   & {\uline{54.8}}    & 53.0          & 50.6     \\
                                & SQA-I \cite{sqa}      & \textbf{78.8} & 67.1            & {\uline{68.2}}    & 65.4          & 65.5     \\
                                & Overall      & \textbf{0.97} & {\uline{0.95}}      & 0.94          & 0.89          & 0.87     \\
\midrule[1pt] 
\multirow{5}{*}{Chart \& OCR}   & ChartQA \cite{chartqa}      & 17.0          & \textbf{17.8}   & {\uline{17.5}}    & 14.6          & 12.6     \\
                                & DocVQA \cite{docvqa}       & 22            & \textbf{23}     & \textbf{23}   & 16            & 8        \\
                                & OCRBench \cite{ocrbench}     & {\uline{30.5}}    & 29.6            & \textbf{32.2} & 21.1          & 2.6      \\
                                & TextVQA \cite{textvqa}     & 57.7          & \textbf{58.1}   & {\uline{57.9}}    & 54.0          & 44.0     \\
                                & Overall      & 0.96          & {\uline{0.98}}      & \textbf{0.99} & 0.78          & 0.47     \\
\midrule[1pt] 
\multirow{4}{*}{Vision-Centric} & MMVP \cite{eyeswideshut}         & 22.7          & \textbf{24.7}   & 20.7          & {\uline{24.0}}    & 9.3      \\
                                & RealWorldQA \cite{realworldqa}  & {\uline{54.8}}    & \textbf{55.8}   & 54.5          & 51.8          & 48.0     \\
                                & CV-Bench \cite{cambrian}     & 54.3          & \textbf{59.6}   & {\uline{58.0}}    & 56.1          & 50.0     \\
                                & Overall      & 0.94          & \textbf{1.00}   & 0.93          & {\uline{0.95}}    & 0.69     \\
\midrule[1pt] 
\multirow{3}{*}{Fine-Grained}   & V*-Bench \cite{vstar}     & 45.6          & {\uline{47.6}}      & 46.6          & \textbf{48.7} & 39.3     \\
                                & HR-Bench \cite{hrbench}    & 35.6          & 36.1            & \textbf{37.3} & {\uline{36.3}}    & 35.2     \\
                                & Overall      & 0.95          & 0.97            & {\uline{0.98}}    & \textbf{0.99} & 0.88     \\
\midrule[1pt]
\multirow{2}{*}{Hallucination}  & POPE   \cite{pope}       & {\uline{86.9}}    & {\uline{86.9}}      & \textbf{87.7} & 85.4          & 78.8     \\
                                & Overall      & {\uline{0.99}}    & {\uline{0.99}}      & \textbf{1.00} & 0.97          & 0.90     \\
\midrule[1pt]
\multicolumn{2}{c|}{Average Overall} & 0.96          & \textbf{0.98}   & {\uline{0.97}}    & 0.91          & 0.76     \\
\bottomrule[2pt]
\end{tabular}}
\end{table*}

\subsection{Enhancing Performance with Layer Combinations}
Building on the strong performance of the penultimate layer, we explored whether combining its features with those from other layers could further enhance task-specific performance. To avoid increasing the number of visual tokens, we adopted a simple concatenation approach along the channel dimension. We trained four additional models: LLaVA-L23\&24, LLaVA-L23\&18, LLaVA-L23\&12, and LLaVA-L23\&6, each combining the 23rd layer’s features with the 24th, 18th, 12th, and 6th layers, respectively.

The results in Table~\ref{combined-layer-visual-features} show that all combinations, except LLaVA-L23\&24, outperformed the model using only the 23rd layer’s features across all task categories, demonstrating the complementary nature of combining visual features from different encoder layers. However, performance still varied across combinations, with no single combination excelling across all tasks. Specifically, LLaVA-L23\&18 achieved the best overall performance, particularly excelling in Knowledge and Vision-Centric tasks. LLaVA-L23\&12 ranked second overall, outperforming in General, Chart \& OCR, and Object Hallucination tasks. On the other hand, LLaVA-L23\&24 showed the least improvement, as the features from the 23rd and 24th layers were highly similar, offering limited information gains. Despite the 6th layer’s poor individual performance, LLaVA-L23\&6 performed best in Fine-Grained tasks, emphasizing the crucial role of low-level features in perceiving intricate visual details. These results highlight the complementary nature of hierarchical visual features in LVLMs, and suggest that task-specific demands are key to determining the most effective feature combinations.

\begin{table*}[t!]
\caption{Performance comparison of LVLMs trained with the penultimate-layer visual features combined with visual features from another layer. The numbers in parentheses represent the score changes relative to the model trained with only penultimate-layer features.``Overall Change" is the average percentage change in performance for each task category. Bold values represent the best score in each row, while underlined values denote the second-best score.}
\label{combined-layer-visual-features}
\renewcommand{\arraystretch}{1}
\centering
\resizebox{\textwidth}{!}{  
\begin{tabular}{l|l|cccc}
\toprule[2pt]
\multicolumn{2}{c|}{\diagbox[innerwidth=5.6cm,innerleftsep=5pt,innerrightsep=5pt]{Benchmarks}{Models}} & LLaVA-L23\&24        & LLaVA-L23\&L18       & LLaVA-L23\&L12          & LLaVA-L23\&L6        \\

\midrule[1pt]
\multirow{5}{*}{General}        & MME-p        & 1464.2 (-16.4)       & {\uline{1495.5 (+14.9)}} & \textbf{1520.0 (+39.4)} & 1431.6 (-49.0)       \\
                                & GQA          & 62.9 (+1.0)          & \textbf{63.4 (+1.5)} & {\uline{63.3 (+1.4)}}       & 63.0 (+1.1)          \\
                                & SEED-I       & 67.2 (+0.8)          & \textbf{68.3 (+1.9)} & {\uline{67.9 (+1.5)}}       & 67.8 (+1.4)          \\
                                & MMB-en       & 65.2 (+1.3)          & \textbf{66.2 (+2.3)} & {\uline{65.8 (+1.9)}}       & 65.3 (+1.4)          \\
                                & Overall Change & +0.94\%          & +2.74\%              & +2.54\%                 & +0.69\%              \\
\midrule[1pt]

\multirow{5}{*}{Knowledge}      & MathVista    & 8.1 (+1.1)           & {\uline{9.5 (+2.5)}}     & \textbf{9.7 (+2.7)}     & 6.4 (-0.6)           \\
                                & MMMU         & 35.7 (+0.8)          & {\uline{36.3 (+1.4)}}    & 36.1 (+1.2)             & \textbf{37.6 (+2.7)} \\
                                & AI2D         & 56.0 (+0.8)          & \textbf{57.0 (+1.8)} & {\uline{56.9 (+1.7)}}       & 55.1 (-0.1)          \\
                                & SQA-I        & 69.5 (+2.4)          & 70.7 (+3.6)          & 68.6 (+1.5)             & 69.0 (+1.9)          \\
                                & Overall Change & +5.76\%          & +12.09\%             & +11.83\%                & +0.45\%              \\
\midrule[1pt] 

\multirow{5}{*}{Chart \& OCR}   & ChartQA      & 17.4 (-0.4)          & \textbf{18.5 (+0.7)} & \textbf{18.5 (+0.7)}    & 18.0 (+0.2)          \\
                                & DocVQA       & \textbf{24 (+1)}     & \textbf{24 (+1)}     & \textbf{24 (+1)}        & \textbf{24 (+1)}     \\
                                & OCRBench     & 32.1 (+2.5)          & 31.9 (+2.3)          & \textbf{32.8 (+3.2)}    & {\uline{32.7 (+3.1)}}    \\
                                & TextVQA      & 58.8 (+0.7)          & \textbf{59.5 (+1.4)} & {\uline{59.2 (+1.1)}}       & 58.3 (+0.2)          \\
                                & Overall Change & +2.94\%          & +4.62\%              & +5.25\%                 & +4.07\%              \\
\midrule[1pt] 

\multirow{4}{*}{Vision-Centric} & MMVP         & 22.0 (-2.7)          & {\uline{26.7 (+2.0)}}    & {\uline{26.7 (+2.0)}}       & \textbf{27.3 (+2.6)} \\
                                & RealWorldQA  & {\uline{55.6 (-0.2)}}    & \textbf{57.0 (+1.2)} & 54.4 (-0.4)             & 54.0 (-1.8)          \\
                                & CV-Bench     & \textbf{60.2 (+0.6)} & 59.2 (-0.4)          & \textbf{60.2 (+0.6)}    & 60.1 (+0.5)          \\
                                & Overall Change & -3.43\%          & +3.19\%              & +2.20\%                 & +2.71\%              \\
\midrule[1pt] 

\multirow{3}{*}{Fine-Grained}   & V*-Bench     & 45.0 (-2.6)          & 48.2 (+0.6)          & {\uline{48.7 (+1.1)}}       & \textbf{50.8 (+3.2)} \\
                                & HR-Bench     & 35.8 (-0.3)          & \textbf{39.1 (+3.0)} & 35.5 (-0.6)             & {\uline{38.6 (+2.5)}}    \\
                                & Overall Change & -3.15\%          & +4.79\%              & +0.32\%                 & +6.82\%              \\
\midrule[1pt]

\multirow{2}{*}{Hallucination}  & POPE         & 87.5 (+0.6)          & {\uline{87.7 (+0.8)}}    & \textbf{87.8 (+0.9)}    & 87.5 (+0.6)          \\
                                & Overall Change & +0.69\%          & +0.92\%              & +1.04\%                 & +0.69\%              \\
\midrule[1pt]

\multicolumn{2}{c|}{Average Overall Change} & +0.62\%              & +4.68\%              & +3.86\%                 & +2.57\%              \\
\bottomrule[2pt]
\end{tabular}}
\end{table*}

\subsection{Optimizing Feature Integration for Task-specific Performance}

To further explore the impact of hierarchical visual features, we explored integrating features from all encoder layers alongside the penultimate layer. The 24 layers of the vision encoder were divided into four distinct visual groups: Low (1st–6th layers), Low-to-mid (7th–12th layers), Mid-to-high (13th–18th layers), and High (19th–24th layers). Within each group, average pooling was applied to generate group-wise representative features, which were then fused using weighted summation. Finally the fused features were concatenated with the 23rd layer’s features along the channel dimension to avoid increasing visual token count. 

We tested five weight distributions across the visual groups: Uniform (0.25, 0.25, 0.25, 0.25), Monotonic increase (0.1, 0.2, 0.3, 0.4), Monotonic decrease (0.4, 0.3, 0.2, 0.1), Increase-then-decrease (0.1, 0.4, 0.4, 0.1), and Decrease-then-increase (0.4, 0.1, 0.1, 0.4), aiming to investigate how varying weight allocation among the groups influence task-specific performance.

As shown in Table~\ref{weighting-configurations}, integrating features from all layers consistently outperformed single-layer models (e.g., Table~\ref{single-layer-visual-features}) across all tasks. However, simply averaging these features (uniform weighting) produced suboptimal results, particularly in Knowledge and Vision-Centric tasks. The Monotonic increase configuration achieved the best overall performance, emphasizing the importance of retaining higher-level features for semantic richness. Despite this, performance variation was still observed at the benchmark level, with each configuration excelling in at least one benchmark. These findings highlight two critical insights: 1) different tasks rely on distinct hierarchical features, necessitating task-aware fusion methods; and 2) static weighting strategies are insufficient, underscoring the need for dynamic adjustment of fusion weights to optimize model performance. 

\begin{table*}[t!]
\caption{Performance comparison of different weighting configurations for multi-layer feature fusion. The overall score is calculated by normalizing the scores for each benchmark (setting the highest score to 1 and adjusting the others accordingly), followed by averaging the normalized scores across all benchmarks within each category. Bold values indicate the best score in each row, while underlined values denote the second-best score.
}
\label{weighting-configurations}
\renewcommand{\arraystretch}{1}
\centering
\resizebox{\textwidth}{!}{  
\begin{tabular}{l|l|ccccc}
\toprule[2pt]
\multicolumn{2}{c|}{\diagbox[innerwidth=5.6cm,innerleftsep=5pt,innerrightsep=5pt]{Benchmarks}{Models}} & Uniform         & Monotonic Increase & Monotonic decrease & Increase-then-decrease & Decrease-then-increase \\ 
\midrule[1pt]
\multirow{5}{*}{General}        & MME-p       & \textbf{1478.7} & {\uline{1477.2}}       & 1465.6             & 1476.2                 & 472.1 \\
                                 & GQA         & 63.3            & \textbf{63.5}      & 63.2               & {\uline{63.4}}             & 63.2   \\
                                 & SEED-I      & \textbf{68.0}   & 67.9               & 67.7               & \textbf{68.0}          & 67.8\\
                                 & MMB-en      & \textbf{65.5}   & 65.3               & \textbf{65.5}      & 65.4                   & \textbf{65.5}\\
                                 & Overall     & \textbf{0.999}  & {\uline{0.998}}        & 0.996              & {\uline{0.998}}            & 0.828\\

\midrule[1pt]

\multirow{5}{*}{Knowledge}      & MathVista   & 7.7             & \textbf{8.4}       & {\uline{8.2}}          & 7.9                    & 8.1\\
                                 & MMMU        & 36.0            & 35.9               & {\uline{36.4}}         & \textbf{36.6}          & 35.9\\
                                 & AI2D        & 56.2            & 56.1               & {\uline{56.3}}         & 56.2                   & \textbf{56.4}\\
                                 & SQA-I       & 69.6            & 69.7               & \textbf{69.8}      & \textbf{69.8}          & \textbf{69.8}\\
                                 & Overall     & 0.973           & \textbf{0.994}     & {\uline{0.992}}        & 0.984                  & 0.986 \\

\midrule[1pt]

\multirow{5}{*}{Chart \& OCR}   & ChartQA     & {\uline{18.9}}      & \textbf{19.2}      & 18.3               & 18.6                   & {\uline{18.9}}\\
                                 & DocVQA      & \textbf{24}     & \textbf{24}        & \textbf{24}        & \textbf{24}            & \textbf{24}  \\
                                 & OCRBench    & {\uline{32.7}}      & 32.6               & 32.4               & \textbf{33.1}          & 32.3   \\
                                 & TextVQA     & 58.8            & {\uline{58.9}}         & \textbf{59.0}      & 58.7                   & 58.8   \\
                                 & Overall     & {\uline{0.992}}     & \textbf{0.996}     & 0.983              & 0.991                  & 0.989  \\

\midrule[1pt]

\multirow{4}{*}{Vision-Centric} & MMVP        & 24.7            & \textbf{28.0}      & {\uline{26.0}}         & {\uline{26.0}}             & 25.3  \\
                                 & RealWorldQA & 56.0            & 55.2               & {\uline{56.2}}         & \textbf{56.5}          & 56.0  \\
                                 & CV-Bench    & 60.6            & \textbf{60.7}      & \textbf{60.7}      & \textbf{60.7}          & 60.5   \\
                                 & Overall     & 0.957           & \textbf{0.992}     & 0.974              & {\uline{0.976}}            & 0.964   \\

\midrule[1pt]

\multirow{3}{*}{Fine-Grained}   & V*-Bench    & 48.7            & {\uline{49.7}}         & 49.2               & 48.2                   & \textbf{49.8}  \\
                                 & HR-Bench    & 35.9            & {\uline{36.0}}         & 35.4               & \textbf{36.4}          & 35.3    \\
                                 & Overall     & 0.989           & \textbf{0.999}     & 0.987              & 0.984                  & {\uline{0.992}} \\

\midrule[1pt]

\multirow{2}{*}{Hallucination}  & POPE        & 87.2            & \textbf{87.7}      & 87.2               & 87.3                   & \textbf{87.7}   \\
                                 & Overall     & 0.998           & \textbf{1.000}     & 0.994              & 0.995                  & \textbf{1.000}  \\

\midrule[1pt]

\multicolumn{2}{c|}{Average Overall}     & 0.984           & \textbf{0.997}     & 0.987              & {\uline{0.988}}            & 0.960  \\

\bottomrule[2pt]
\end{tabular}}
\end{table*}

%% file: methods.tex
\section{Methods}

Building on the insights from Section \ref{sec2}, we propose a novel method that dynamically integrates hierarchical visual features based on task-specific instructions. Our method introduces an instruction-guided vision aggregator into the LVLM framework, addressing the limitations of static or task-agnostic fusion in existing methods.

\begin{figure}[t]
  \centering
  \includegraphics[width=\textwidth]{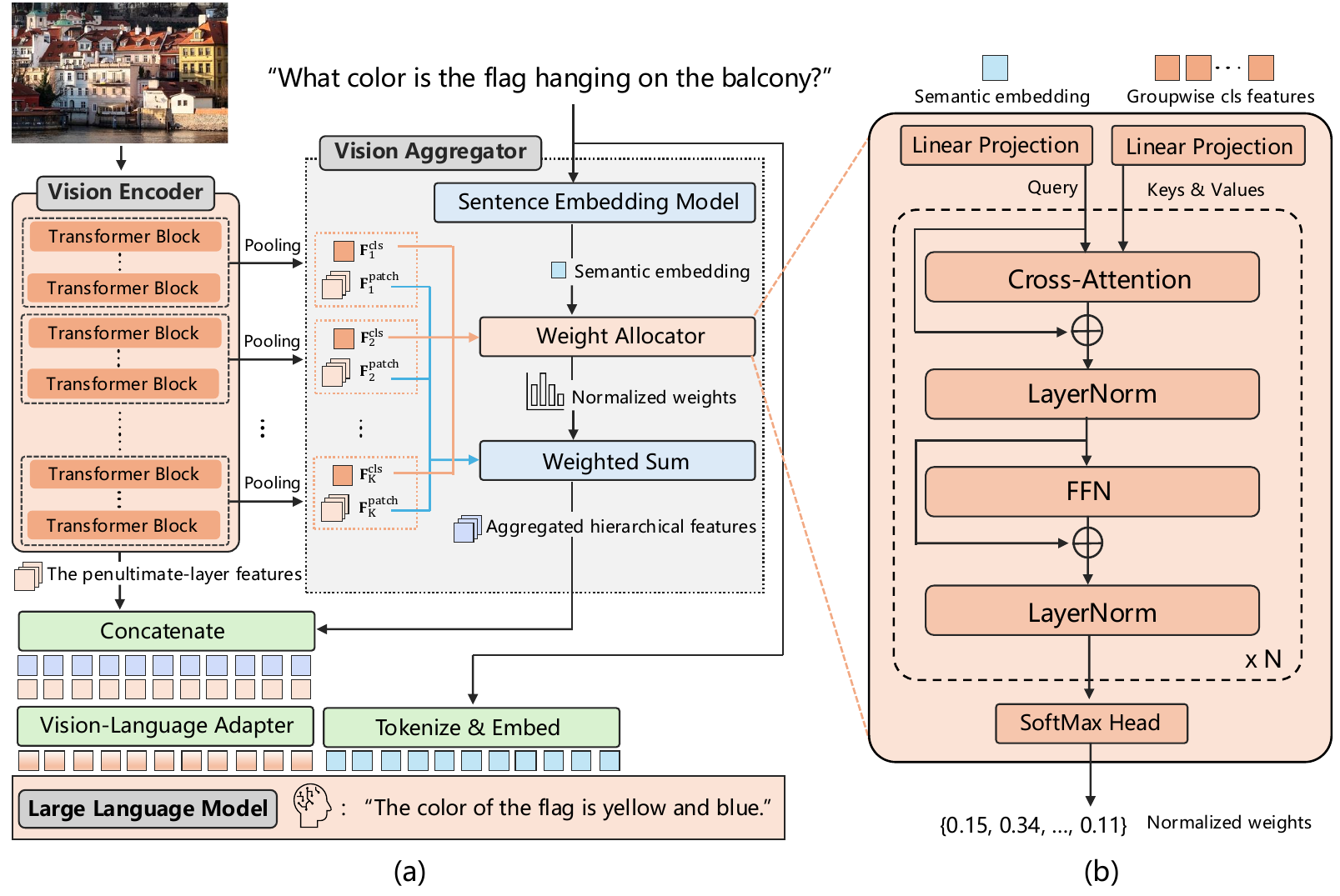}
  \caption{(a) Overview of the proposed framework. (b) Detailed architecture of the weight allocator within the instruction-guided vision aggregator.}
  \label{fig:framework}
\end{figure}

\subsection{Overall Framework}
The overall framework is illustrated in Fig~\ref{fig:framework}(a). It consists of four main modules: a vision encoder $V$, an instruction-guided vision aggregator $\mathit{IGVA}$, a vision-language adapter $ADP$, and a large language model $LLM$. For demonstration purposes, we use the widely-adopted LLaVA-v1.5 \cite{llava1.5} as the implementation framework. Below we introduce the design of each component in detail.

\textbf{Vision Encoder}. We use a CLIP-ViT \cite{clip} as the vision encoder. It divides the input image $\textbf{I} \in \mathbb{R}^{C \times H \times W}$ into small patches and embeds them through a convolution layer. The patch embeddings are then passed through a series of transformer layers to extract visual features. The hierarchical output of the vision encoder is:
\begin{equation}
\textbf{F} = V(\textbf{I}) \in \mathbb{R} ^ {L \times (N+1) \times D},
\end{equation}
where $L$ is the number of transformer layers, $N$ is the number of patches, and $D$ is the dimension of the hidden state. Each layer produces $N+1$ representations, as the CLIP-ViT appends a special $<cls>$ token to the patch embeddings to obtain a global representation.

\textbf{Instruction-guided Vision Aggregator}. The key innovation of our method is the vision aggregator, which dynamically integrates hierarchical visual features based on task-specific instructions. We divide the $L$ transformer layers of the vision encoder into $K$ visual groups, each containing $L/K$ consecutive layers. For each group, average pooling is applied to the $<cls>$ representation across the layers to obtain a group-specific global feature: 
\begin{equation}
\textbf{F}^{cls}_{k} = \mathit{Avg}(\textbf{cls}_{k_1}, \textbf{cls}_{k_2}, ..., \textbf{cls}_{k_{L/K}}) \in \mathbb{R}^{ D}, for \ k = 1, 2, ..., K,
\end{equation}
where $\textbf{cls}_{k_i} \in \mathbb{R}^{D}$ is the  $<cls>$ feature of the i-th layer in group $k$. The aggregator comprises a sentence embedding model and a weight allocator. The sentence embedding model encodes the input instruction into a semantic embedding $\mathbf{s}$, which is used by the weight allocator to compute the importance of each visual group:
\begin{equation}
\textbf{w} = \mathit{IGVA}(\textbf{s}, \textbf{F}^{cls}_{1}, \textbf{F}^{cls}_{2}, ..., \textbf{F}^{cls}_{K}) \in \mathbb{R}^{K}, \quad where \sum_{k=1}^K w_k = 1.
\end{equation}
For each group, we apply average pooling to the patch features. Then we perform weighted summation across the pooled features using the weight vector $\mathbf{w}$: 
\begin{equation}
\textbf{F}^{patch}_k = Avg(\textbf{F}^{patch}_{k_1}, \textbf{F}^{patch}_{k_2}, ..., \textbf{F}^{patch}_{k_{L/K}}) \in \mathbb{R} ^ {N \times D}, \quad for \ k = 1, ..., K .
\end{equation}
\begin{equation}
\textbf{F}_{fused} = \sum_{k=1}^K w_k\times \textbf{F}^{patch}_k \in \mathbb{R} ^ {N \times D},
\end{equation}
where $\textbf{F}^{patch}_{k_i} \in \mathbb{R}^{N\times D}$represents the patch features of the i-th layer in group $k$. Finally, to preserve semantic richness, the fused features are concatenated with the penultimate layer’s features to form the final visual representation:
\begin{equation}
\hat{\textbf{F}} = Concate(\textbf{F}_{fused}, \textbf{F}_{penultimate}) \in \mathbb{R} ^ {N \times 2D}.
\end{equation}

\textbf{Vision-Language Adapter}. The vision-language adapter consists of a two-layer MLP with a GELU activation function \cite{gelu}. It aligns the visual features with the input space of the LLM:
\begin{equation}
 ADP(\hat{\textbf{F}}) = Proj_2(GELU(Proj_1(\hat{\textbf{F}})) )\in \mathbb{R} ^ {N \times D_t},
\end{equation}
where $D_t$ is the hidden dimension of the LLM.

\textbf{Large Language Model}. The large language model tokenizes the input instruction and computes the embeddings for each token. These embeddings are then concatenated with the visual features along the sequence dimension, forming a combined sequence that is processed by the transformer layers of the LLM to generate a textual response.

\subsection{Details of Instruction-guided Vision Aggregator}
The instruction-guided vision aggregator comprises a sentence embedding model and a weight allocator, working in tandem to facilitate dynamic, task-specific fusion of hierarchical visual features. 

\textbf{Sentence Embedding Model}. The sentence embedding model is implemented using a pre-trained MPNet model \cite{mpnet}. It encodes the task-specific instruction into a semantic vector $\mathbf{s}$, which captures critical dependencies between the instruction and the hierarchical visual features, allowing the weight allocator to effectively prioritize the visual groups.

\textbf{Weight Allocator}. The weight allocator computes dynamic weights for each visual group based on the sentence embedding. As shown in Fig~\ref{fig:framework}(b), it consists of two linear projections, a stack of transformer blocks, and a softmax network. Each transformer block includes a cross-attention module and a feed-forward network, with residual connection and layer normalization applied after each module. The sentence embedding and visual group representations are first transformed through linear projections to match the hidden dimension of the transformer blocks. The projected sentence embedding then serves as the query, while the projected visual features act as keys and values. The output of the transformer blocks is passed through the softmax head to generate a weight vector $\mathbf{w}$, which is used to adaptively sum the features across the visual groups.

%% file: experiments.tex
\section{Experiments}
\subsection{Model Implementation}
We adopt LLaVA-v1.5 \cite{llava1.5} as the implementation framework. It employs a CLIP-ViT-L-336px \cite{clip} as the vision encoder, a two-layer MLP as the vision-language adapter, and a Vicuna-v1.5-7B \cite{vicuna} as the LLM backbone. For the instruction-guided vision aggregator, we employ all-mpnet-base-v2 \cite{mpnet} as the sentence embedding model. The weight allocator comprises 4 cross-attention-based transformer blocks, each with a hidden dimension of 1024 and 4 attention heads.

We partition the 24 layers of the vision encoder into four distinct visual groups: layers 1 to 6 represent low-level features, layers 7 to 12 represent low-to-mid-level features, layers 13 to 18 represent mid-to-high-level features, and layers 19 to 24 represent high-level features. 

\subsection{Training Scheme}
We follow the two-stage training strategy of LLaVA-v1.5 \cite{llava1.5}:

In the pre-training stage, we leverage the LLaVA-LCS-558K dataset, which consists of 558K image-text pairs. Since no instruction data is available at this stage, the instruction-guided vision aggregator is not employed. Instead, features from the four visual groups are fused via average pooling. During this phase, only the adapter is updated, while the vision encoder and LLM remain frozen.

In the instruction-tuning stage, we utilize the LLaVA-Instruct-665K dataset, which contains 665K multimodal instruction-tuning samples. The proposed vision aggregator is fully integrated to the architecture in this stage. The weight allocator, adapter, and LLM are jointly trained, while the vision encoder and sentence embedding model remain frozen.

During instruction-tuning, we observed a tendency for the weight allocator to assign disproportionally high weights to the mid-to-high-level visual group while neglecting others. Therefore we introduce an entropy-based auxiliary loss function for weight balancing:
\begin{equation}
    L_{total} = L_{lm} + \lambda \times \sum_{k=1}^K w_klog(w_k),
\end{equation}
where $L_{lm}$ is the original language modeling loss, $w_k$ represents the weight assigned to the k-th visual group, and $\lambda$ is a regularization coefficient set to 0.02.

The training hyperparameters are also inherited from LLaVA-v1.5. The optimizer is AdamW \cite{adamw} with a batch size of 256 for pre-training and 128 for instruction-tuning. The learning rate is 1e-3 for pretraining and 2e-5 for instruction-tuning, both following a cosine decay schedule with a 0.03 warmup ratio and no weight decay. 

\subsection{Comparison with Baseline and Existing Fusion Methods} 
Table~\ref{model-comparison-transposed} presents a comparison of our method with the baseline model and two existing multi-layer visual feature fusion approaches across 10 benchmarks. These benchmarks include traditional VQA datasets (GQA \cite{gqa}, SQA \cite{sqa}, TextVQA \cite{textvqa}, VizWiz \cite{vizwiz}) and LVLM-specific benchmarks (MMB-en \cite{mmb}, MM-Vet \cite{mmvet}, SEED-I \cite{seed}, MMMU \cite{mmmu}, MME-P \cite{mme}, POPE \cite{pope}). 

\begin{table*}[t!]
\caption{Performance comparison of our method against the baseline model (LLaVA-v1.5 \cite{llava1.5}) and existing hierarchical visual feature fusion methods (DenseConnector \cite{denseconnector} and MMFuser \cite{mmfuser}) across 4 traditional VQA benchmarks and 6 LVLM-specific benchmarks. All fusion methods are implemented on the LLaVA-v1.5 framework. Bold values indicate the best score in each row, while underlined values represent the second-best score.}
\label{model-comparison-transposed}
\renewcommand{\arraystretch}{1.5}
\centering
\resizebox{\textwidth}{!}{
\begin{tabular}{l|cccccccccc}
\toprule[2pt]
{\diagbox[innerleftsep=8pt,innerrightsep=16pt]{Benchmarks}{Models}}       & LLaVA-v1.5-7B         & DenseConnector   &  MMFuser           &  Ours             \\
\midrule[1pt]
\hline
GQA \cite{gqa}                     & 61.9                        & \textbf{63.8}               & 62.8                        & \uline{63.1}                \\
SQA \cite{sqa}                     & 67.1                        & \uline{69.5}                & 68.7                        & \textbf{70.2}               \\
TextVQA \cite{textvqa}             & 58.1                        & \uline{59.2}                & 58.8                        & \textbf{59.4}               \\
VizWiz \cite{vizwiz}               & 53.2                        & -                           & \uline{53.4}                 & \textbf{54.3}                \\
\hline
MMB-en \cite{mmb}                  & 63.9                        & 66.8                        & \textbf{67.5}               & \uline{66.9}                \\
MM-Vet \cite{mmvet}                & \uline{32.8}                 & 32.7                        & 32.6                        & \textbf{33.5}               \\
SEED-I \cite{seed}                 & \uline{67.2}                 & -                           & 60.8                        & \textbf{68.3}               \\
MMMU \cite{mmmu}                   & \uline{34.9}                 & 34.8                        & -                           & \textbf{36.4}               \\
MME-p \cite{mme}                   & 1480.6                      & -                           & 1479.7                      & \textbf{1519.8}              \\
POPE \cite{pope}                   & 86.9                        & 86.6                        & 86.3                        & \textbf{87.8}               \\
\bottomrule[2pt]
\end{tabular}}
\end{table*}

On traditional VQA benchmarks, our method outperformed the baseline model and the two competing fusion methods in most cases.  It consistently achieved the highest scores on SQA, TextVQA, and VizWiz, demonstrating robustness across various tasks. While DenseConnector slightly surpassed our approach on GQA, we still outperformed both the baseline and MMFuser, showcasing competitive performance even on challenging datasets.

For LVLM-specific benchmarks, our method excelled, securing the highest scores on SEED-I, MMMU, MME-P, and POPE. It ranked second on MMB-en, trailing DenseConnector but significantly outperforming MMFuser and the baseline. These results highlight the ability of our method to effectively handle the diverse and complex task instructions for LVLMs.

Overall, our method achieved the best score on 8 out of the 10 benchmarks, validating the effectiveness of instruction-guided fusion of hierarchical visual features. This confirms that task-specific adaptation of visual features enhances LVLM performance, particularly in tasks requiring complex visual-textual interactions.

\subsection{Comparison with Popular 7B-Scale LVLMs}
Table~\ref{7B-comparison} compares our method with other popular 7B-scale LVLMs across traditional VQA and LVLM-specific benchmarks. Despite using less training data and smaller image resolutions, our model consistently achieved the highest or second-highest scores across all benchmarks.

\begin{table*}[t!]
\caption{Performance comparison of our method with widely-used 7B-scale LVLMs across 4 traditional VQA benchmarks and 6 LVLM-specific benchmarks. For each model, it reports the performance score, LLM backbone, resolution constraints, and the amount of training data used. Bold values represent the best score in each row, while underlined values denote the second-best score.} 
\label{7B-comparison}
\renewcommand{\arraystretch}{1.5}
\centering
\resizebox{\textwidth}{!}{
\setlength{\tabcolsep}{0.5mm}{
\begin{tabular}{l|c|c|c|c|c|c|c|c|c|c|c|c|c}
\toprule[2pt]
Models & \makecell{LLaVA-v1.5 \\ \cite{llava1.5}} & \makecell{InstructBLIP \\ \cite{instructblip}} & \makecell{mPLUG-Owl2 \\ \cite{mplugowl2}} & \makecell{MiniGPT-v2 \\ \cite{minigptv2}} & \makecell{LLaMA-Adapter-v2 \\ \cite{llamaadapterv2}} & \makecell{IDEFICS \\ \cite{idefics}}  & \makecell{Qwen-VL-Chat \\ \cite{qwenvl}} & Ours \\
\midrule[1pt]
\midrule[1pt]
LLM Backbone & Vicuna-v1.5-7B & Vicuna-7B & LLaMA 2-7B & LLaMA 2-7B & LLaMA-7B & LLaMA-7B & Qwen-7B  & Vicuna-v1.5-7B \\
Resolution & 336 & 224 & 448 & 448 & 336 & 224 & 448 & 336 \\
Training Data & 0.5M+0.6M & 129M+14.7M & 384M+1.2M & - & 0.6M & 1.6B & 1.4B+50M+0.3M & 0.5M+0.6M \\
\midrule
GQA \cite{gqa} & \uline{61.9} & 49.2 & 56.1 & 60.1 & - & 38.4  & 57.5 & \textbf{63.1} \\

SQA \cite{sqa} & 67.1 & 60.5 & \uline{68.7} & - & - & -  & 68.2 & \textbf{70.2} \\
TextVQA \cite{textvqa} & 58.1 & 50.1 & 54.3 & - & - & 25.9  & \textbf{61.5} & \uline{59.4} \\
VizWiz \cite{vizwiz} & 53.2 & 34.5 & \textbf{54.5} & 53.6 & - & 35.5 & 38.9 & \uline{54.3} \\
\midrule
MMB-en \cite{mmb} & 63.9 & 33.9 & \uline{64.5} & 9.4 & 41.0 & 48.2 & 60.6 & \textbf{66.9} \\
MM-Vet \cite{mmvet} & 32.8 & 26.2 & \textbf{36.2} & - & 31.5 & - & -  & \uline{33.5} \\
SEED-I \cite{seed} & \uline{67.2} & 58.8 & 57.8 & - & 32.7 & -  & 65.4 & \textbf{68.3} \\
MMMU \cite{mmmu} & 34.9 & - & - & - & 29.8 & - & \uline{35.9} & \textbf{36.4} \\
MME-p \cite{mme} & \uline{1480.6} & - & 1450.2 & - & 972.7 & -  & 1487.6 & \textbf{1519.8} \\
POPE \cite{pope} & \uline{86.9} & - & 86.2 & - & - & - & -  & \textbf{87.8} \\
\bottomrule[2pt]
\end{tabular}}}
\end{table*}

On traditional VQA tasks, our model excelled by delivering top performance on SQA and GQA, while achieving the second-best results on TextVQA and VizWiz. It consistently outperformed the baseline model and matched or exceeded the performance of models trained on significantly larger datasets, such as Qwen-VL-Chat \cite{qwenvl} and mPLUG-Owl2 \cite{mplugowl2}, showcasing the power of insturction-guided hierarchical feature fusion even with limited instruction-tuning data.

For LVLM-specific benchmarks, our model achieved the highest scores on MMB-en, SEED-I, MMMU, MME-P, and POPE. Although mPLUG-Owl2 \cite{mplugowl2}, with a higher image resolution and larger training data, achieved stronger results on MM-Vet, it underperformed our method on other benchmarks, revealing a lack of consistency compared to our approach. These results emphasize the balanced adaptability of our method in handling both general and specialized tasks.

Overall, the results validate the effectiveness of our method. By dynamically adapting to task-specific requirements, it  demonstrates consistent strengths across both traditional VQA tasks and comprehensive LVLM-specific benchmarks, outperforming or matching the competitors with greater consistency and data efficiency.

\subsection{Ablation Study}
We conducted ablation experiments to assess the contribution of each component in our method, as summarized in Table~\ref{ablation-comparison}. Starting with the full configuration, we progressively removed key components to analyze their impact on performance.

\begin{table*}[t!]
\caption{Ablation study results. Starting with the complete configuration, key components are progressively removed: (1) removal of the weight balancing loss, (2) replacement of the instruction-guided vision aggregator with an average pooling mechanism, and (3) elimination of multi-level visual feature utilization, reverting to the baseline model.}
\label{ablation-comparison}
\renewcommand{\arraystretch}{1.5}
\centering
\resizebox{\textwidth}{!}{
\begin{tabular}{l|l|cccc}
\toprule[2pt]
\multicolumn{2}{c|}{\diagbox[innerwidth=5.6cm,innerleftsep=5pt,innerrightsep=5pt]{Benchmarks}{Models}} & Our method & w/o auxiliary loss & w/o aggregator & w/o multi-layer features \\
\midrule[1pt]
\multirow{4}{*}{General}        & MME-p \cite{mme}       & 1519.8 & 1508.7             & 1478.7         & 1480.6                   \\
                                 & GQA \cite{gqa}         & 63.1            & 62.9               & 63.3  & 61.9                     \\
                                 & SEED-I \cite{seed}      & 68.3   & 68.5               & 68.0           & 66.4                     \\
                                 & MMB-en \cite{mmb}     & 66.9            & 66.5               & 65.5           & 63.9                     \\
\midrule[1pt]
\multirow{4}{*}{Knowledge}      & MathVista \cite{mathvista}   & 10.5            & 9.5                & 7.7            & 7.0                      \\
                                 & MMMU \cite{mmmu}       & 36.4   & 35.6               & 36.0           & 34.9                     \\
                                 & AI2D \cite{ai2d}        & 57.0            & 56.9               & 56.2           & 55.2                     \\
                                 & SQA-I \cite{sqa}      & 70.2   & 70.0               & 69.6           & 67.1                     \\
\midrule[1pt]
\multirow{3}{*}{Chart \& OCR}   & ChartQA \cite{chartqa}     & 18.3            & 17.9               & 18.9  & 17.8                     \\
                                 & OCRBench \cite{ocrbench}    & 33.4            & 32.9               & 32.7           & 29.6                     \\
                                 & TextVQA \cite{textvqa}     & 59.4   & 59.2               & 58.8           & 58.1                     \\
\midrule[1pt]
\multirow{3}{*}{Vision-Centric} & MMVP \cite{eyeswideshut}        & 25.3            & 25.0               & 24.7           & 24.7                     \\
                                 & RealWorldQA \cite{realworldqa} & 48.0            & 56.4               & 56.0           & 55.8                     \\
                                 & CV-Bench \cite{cambrian}    & 61.7            & 61.2               & 60.6           & 59.6                     \\
\midrule[1pt]
\multirow{2}{*}{Fine-Grained}   & V*-Bench \cite{vstar}    & 48.2            & 48.0               & 48.7           & 47.6                     \\
                                 & HR-Bench \cite{hrbench}    & 40.0            & 37.5               & 35.9           & 36.1                     \\
\midrule[1pt]
Hallucination                   & POPE \cite{pope}        & 87.8   & 87.8               & 87.2           & 86.9                     \\
\bottomrule[2pt]
\end{tabular}}
\end{table*}

First, removing the weight balancing loss led to moderate performance drops across most benchmarks, such as a decrease from 1519.8 to 1508.7 on MME-p and from 40.0 to 38.5 on HR-bench. The auxiliary loss helps prevent over-reliance on specific visual groups, ensuring a more balanced contribution from each group, which is crucial for overall performance.

Next, replacing the instruction-guided vision aggregator with a simple average pooling mechanism for feature fusion resulted in more significant performance declines. Tasks requiring fine-grained understanding or complex reasoning saw drops, such as from 38.5 to 35.9 on HR-bench and from 9.5 to 7.7 on MathVista. This highlights the importance of the instruction-guided mechanism in dynamically assigning weights to hierarchical features based on task needs.

Finally, removing multi-level feature fusion, which reverted the model to using only the penultimate layer's features, caused the largest performance declines. The SEED-I score dropped from 68.0 to 66.4, and SQA decreased from 69.6 to 67.1. These results underscore the critical role of multi-level features in enhancing the model's visual perception by integrating complementary information from different layers.

Overall, the ablation study confirms the necessity of each component in our method. The load balancing loss ensures a balanced contribution from all visual groups, the instruction-guided vision aggregator facilitates task-specific adaptation, and multi-level feature fusion enables the model to leverage complementary visual information from different encoder layers. They together form an integrated system that drives the superior performance of our method.

\subsection{Insights from weight allocation}

Table~\ref{Average weight} presents the average weight distributions assigned by the trained vision aggregator to the four visual groups across various downstream tasks. These distributions confirm the task-specific dependencies on hierarchical visual features and validate the effectiveness of the instruction-guided mechanism.

\begin{table*}[t!]
\caption{Average weight distributions assigned by the instruction-guided vision aggregator across four hierarchical visual groups for each benchmark and task category.}
\label{Average weight}
\renewcommand{\arraystretch}{0.9}
\centering
\resizebox{0.8\textwidth}{!}{
\begin{tabular}{l|l|cccc}
\toprule[2pt]
\multicolumn{2}{c|}{\diagbox[innerwidth=5.6cm,innerleftsep=5pt,innerrightsep=5pt]{Benchmarks}{Visual groups}} & low & low-to-mid & mid-to-high & high \\
\midrule[1pt]
\multirow{5}{*}{General}        & MME-p \cite{mme}       & 0.16 & 0.27       & 0.35        & 0.21 \\
                                 & GQA \cite{gqa}        & 0.19 & 0.19       & 0.34        & 0.22 \\
                                 & SEED-I \cite{seed}      & 0.15 & 0.22       & 0.35        & 0.27 \\
                                 & MMB-en \cite{mmb}     & 0.16 & 0.23       & 0.34        & 0.27 \\
                                 & Overall     & 0.17 & 0.23       & 0.35        & 0.24 \\
\midrule[1pt]
\multirow{5}{*}{Knowledge}      & MathVista \cite{mathvista}   & 0.19 & 0.31       & 0.31        & 0.19 \\
                                 & MMMU \cite{mmmu}        & 0.16 & 0.28       & 0.35        & 0.21 \\
                                 & AI2D \cite{ai2d}        & 0.17 & 0.29       & 0.34        & 0.20 \\
                                 & SQA-I \cite{sqa}       & 0.18 & 0.30       & 0.32        & 0.20 \\
                                 & Overall     & 0.18 & 0.30       & 0.33        & 0.20 \\
\midrule[1pt]
\multirow{4}{*}{Chart \& OCR}   & ChartQA \cite{chartqa}     & 0.13 & 0.29       & 0.41        & 0.17 \\
                                 & OCRBench \cite{ocrbench}    & 0.14 & 0.28       & 0.38        & 0.20 \\
                                 & TextVQA \cite{textvqa}     & 0.15 & 0.25       & 0.38        & 0.22 \\
                                 & Overall     & 0.14 & 0.27       & 0.39        & 0.20 \\
\midrule[1pt]
\multirow{4}{*}{Vision-Centric} & MMVP \cite{eyeswideshut}        & 0.17 & 0.26       & 0.33        & 0.24 \\
                                 & RealWorldQA \cite{realworldqa} & 0.23 & 0.30       & 0.28        & 0.19 \\
                                 & CV-Bench \cite{cambrian}    & 0.20 & 0.29       & 0.31        & 0.20 \\
                                 & Overall     & 0.20 & 0.28       & 0.31        & 0.21 \\
\midrule[1pt]
\multirow{3}{*}{Fine-Grained}   & V*-Bench \cite{vstar}    & 0.23 & 0.29       & 0.29        & 0.19 \\
                                 & HR-Bench \cite{hrbench}    & 0.17 & 0.26       & 0.36        & 0.21 \\
                                 & Overall     & 0.20 & 0.28       & 0.33        & 0.20 \\
\midrule[1pt]
Hallucination                   & POPE \cite{pope}        & 0.18 & 0.25       & 0.35        & 0.22 \\
\bottomrule[2pt]
\end{tabular}}
\end{table*}

For General tasks, the mid-to-high visual group is most heavily weighted, with moderate support from the low-to-mid and high groups. This indicates that general VQA tasks primarily rely on mid- and high-level visual features. 
In Knowledge tasks, both the low-to-mid and mid-to-high groups receive high weights, suggesting a balanced dependency on different levels of visual features. Interestingly, the high group has relatively lower weights in these reasoning-heavy tasks, indicating limited reliance on global features for such scenarios. 
For Chart \& OCR tasks, the mid-to-high group are the most influential, highlighting its role in extracting structured visual information. 
In Vision-Centric tasks, the weights are more evenly distributed, reflecting the need for both global and localized visual understanding.
In Fine-Grained tasks, the mid-to-high group again leads, but the low group sees a notable weight increase, demonstrating the significance of low-level features for capturing intricate visual details. 
For Object Hallucination tasks, the mid-to-high group achieves the highest weight, showcasing its importance in object-level perception, while the low-to-mid group provides complementary support to improve accuracy.

In summary, mid-to-high-level features involves the most versatile and critical visual information, particularly excelling in tasks that require semantic understanding and structured information processing. Low-level features, while contributing less overall, are indispensable in several Fine-Grained and Vision-Centric tasks, where capturing intricate visual details is essential. Low-to-mid features play a key role in reasoning-heavy tasks, such as Knowledge benchmarks, by providing crucial contextual information. High-level features serve more as auxiliary support, primarily aiding in global semantic understanding for tasks like Vision-Centric and General benchmarks. These observations demonstrate the effectiveness of our dynamic fusion strategy in capturing task-specific dependencies and leveraging the complementary strengths of hierarchical visual features.

%% file: conclusion.tex
\section{Conclusion}
In this paper, we analyzed the task dependencies of LVLMs on different levels of visual features and introduced the instruction-guided vision aggregator, a module that adaptively fuses hierarchical visual features based on task-specific instructions. Integrated into the LLaVA-v1.5 framework, our method achieved significant improvements over the baseline, outperforming existing fusion methods and similarly scaled LVLMs.

While our method shows promising results, some limitations remain. It relies on clear and high-quality textual instructions, which may pose challenges in cases of ambiguous or complex task descriptions. Furthermore, although the method avoids increasing the number of visual tokens, integrating an additional module introduces minor computational overhead. Future work will focus on addressing these limitations to further enhance the generalization and efficiency of the approach.